\documentclass[10pt,twocolumn,letterpaper]{article}

\usepackage{cvpr}
\usepackage{multirow}
\usepackage{xcolor}
\usepackage{graphicx}
\usepackage{subcaption}
\usepackage{microtype}

\newcommand{\method}{CondI\xspace}
\definecolor{cvprblue}{rgb}{0.21,0.49,0.74}
\usepackage[pagebackref,breaklinks,colorlinks,allcolors=cvprblue]{hyperref}

\def\paperID{*****} 
\def\confName{CVPR}
\def\confYear{2026}

\title{Conditional Imputation for Within-Modality Missingness \\ in Multi-Modal Federated Learning}

\author{
Wugeng Zheng\thanks{Equal contribution.} \\
University of Central Florida \\
\texttt{wgzheng@ucf.edu}
\and
Ziwen Kan\footnotemark[1] \\
University of Central Florida \\
\texttt{zi605672@ucf.edu}
\and
Katie Wang \\
Independent Researcher \\
\texttt{katiewang23@outlook.com}
\and
Chen Chen \\
University of Central Florida \\
\texttt{chen.chen@crcv.ucf.edu}
\and
Song Wang\thanks{Corresponding author.} \\
University of Central Florida \\
\texttt{song.wang@ucf.edu}
}

\begin{document}
\maketitle
\begin{abstract}

Multimodal Federated Learning (MMFL) enables privacy-preserving collaborative training, but real-world clinical applications often suffer from within-modality missingness caused by sensor intermittency or irregular sampling. Existing methods implicitly represent unobserved data via architectural alignment or missing embeddings, often failing to recover the true distribution and yielding sub-optimal performance. We propose \method, a federated framework explicitly addressing this missingness using conditional diffusion models. \method employs a two-phase training pipeline: first, imputing unobserved temporal components using available multimodal context and conditional embeddings; second, optimizing modality-specific extractors and joint embedding spaces. During inference, imputed raw data pass through trained extractors to generate robust features, providing a holistic representation for downstream tasks. Explicit data imputation ensures models operate on complete semantic structures, significantly enhancing resilience against severe data incompleteness. Experiments on three clinical datasets (PTB-XL, SLEEP-EDF, MIMIC-IV) demonstrate \method achieves comparable results to state-of-the-art baselines. Code: https://github.com/ZhengWugeng/CondI
\end{abstract}

\section{Introduction}

Real-world data are inherently multimodal and often distributed across different clients, making multimodal federated learning (MMFL) an important paradigm \cite{zhao2022multimodal,huang2024multimodal,che2023multimodal} for collaboratively learning from heterogeneous multimodal data while preserving data privacy.
Existing MMFL studies typically assume that all clients share the same set of modalities. 
However, this assumption rarely holds in real-world scenarios. 
For example, in clinical settings such as MIMIC-IV \cite{johnson2020mimiciv}, data missingness is ubiquitous. While extensive research has addressed the absence of entire examinations (i.e., modality missing) \cite{chen2022fedmsplit,phung2024contrastive}, real-world deployments frequently face an additional, finer-grained layer of complexity: within-modality missingness \cite{yu2024robust, nguyen2024fedmac}. Even when monitoring sensors are successfully deployed, irregular sampling intervals or intermittent sensor detachments inevitably fracture the temporal dependencies of physiological signals.
Such fragmented observations introduce significant challenges for effective multimodal representation learning in federated settings.

Missing data in multimodal systems, whether at the modality or within-modality level, often lead to degraded representations if not properly aligned \cite{che2023multimodal}.
When critical temporal segments are fractured, models struggle to maintain semantic continuity, inevitably leading to degraded expressive power.
Handling missingness is further complicated by the privacy-preserving nature of federated data sources.
Importantly, existing solutions for recovering missing modalities, including data imputation \cite{zhou2022missing,chen2024probabilistic,wang2023distribution,yang2024frequency} and generation using pre-trained multimodal foundation models \cite{das2024decoder,goswami2024moment,cao2023tempo,rasul2023lag}, typically assume centralized data access or comprehensive modality coverage, which conflicts with federated learning settings \cite{nguyen2025pepsy}.

\begin{table}[!t]
\centering
\caption{Comparison of state-of-the-art multimodal federated learning frameworks. \method uniquely addresses within-modality missingness by explicitly conditional imputation.}
\label{tab:missing_comparison}
\footnotesize
\setlength{\tabcolsep}{1.5pt} 
\begin{tabular}{lcc}
\toprule
\textbf{Method} & \textbf{Modality-Level} & \textbf{Within-Modality Mechanism} \\
\midrule
FedMSplit \cite{chen2022fedmsplit} & \checkmark  & $\times$ \\
MIFL \cite{phung2024contrastive}   & \checkmark  & $\times$ \\
FedInMM \cite{yu2024robust}        & \checkmark  & Implicit (Feature Alignment) \\
FedMAC \cite{nguyen2024fedmac}     & \checkmark  & Implicit (Alignment / Contrastive Reg.) \\
PEPSY \cite{nguyen2025pepsy}       & \checkmark  & Implicit (Missing Embedding) \\
\textbf{\method (Ours)}            & \textbf{\checkmark} & \textbf{Explicit (Conditional Imputation)} \\
\bottomrule
\end{tabular}%

\end{table}

To mitigate modality absence in federated settings, early methods focused on learning universal representations \cite{chen2024feddat,peng2024fedmm,phung2025federated,xiong2022unified,yu2023multimodal}, yet they heavily relied on the unrealistic assumption of uniform modality configurations across all clients.
More recent FL-specific approaches attempt to address data missingness, but they generally fall into two sub-optimal paradigms. 
First, several frameworks \cite{chen2022fedmsplit, phung2024contrastive} address only modality-level missingness.
Second, methods that consider within-modality missingness, such as PEPSY \cite{nguyen2025pepsy} and others \cite{yu2024robust, nguyen2024fedmac}, typically resort to implicit modeling. They rely on passive architectural alignments or learnable ``missing embeddings'' that merely flag the unobserved data. This passive approach is fundamentally inadequate for clinical applications.
For example, a sudden blood pressure (BP) drop typically triggers a compensatory heart rate (HR) rise \cite{zhang2001dynamic}. Implicit methods merely `flag' missing BP via embeddings, failing to capture critical signal correlations. As noted in \cite{lincyin2026}, missing data may interrupt the multimodal fusion space and the model may overfit on the remaining observed data, introducing semantic ambiguity and degrading expressive power.



To break this expressive bottleneck, we propose to explicitly generate missing observations conditioned on available information. 
Implicit latent-space methods suffer from representation degradation and fail to maintain semantic coherence under severe missingness. In contrast, explicitly restoring raw data preserves the underlying physiological distribution and ensures the generated signals remain clinically verifiable.
Therefore, our framework operates in two phases.
During Phase A, the diffusion generation is explicitly guided by available cross-modal context and a dynamically learned conditional embedding ($\mathcal{W}_{\text{cond}}$) that encodes complex missing patterns, enabling the explicit generation of imputed raw data. 
In Phase B, the framework focuses on downstream task optimization by training dedicated feature extractors to capture instance-level features ($\mathcal{W}_{\text{ins}}$), learning structural modality embeddings ($\mathcal{W}_{\text{mod}}$), and systematically fusing them with the conditional embeddings ($\mathcal{W}_{\text{cond}}$) prior to classification. 
Importantly, during the inference stage, the explicitly imputed raw data are leveraged by these trained extractors to derive continuous instance-level semantics ($\mathcal{W}_{\text{ins}}$). 
Ultimately, as contrasted in Table~\ref{tab:missing_comparison}, our approach distinguishes itself by shifting from passive alignment to explicit data imputation
, which rigorously preserves semantic consistency, enables richer modeling of missing-data distributions, and achieves superior multimodal alignment in federated environments.
Our contributions are summarized as follows:

\begin{itemize}

\item We study the critical problem of within-modality missingness in the multimodal federated learning setting.

\item We propose a new conditional generative diffusion framework to address within-modality missingness issues in multimodal federated learning.


\item Extensive experiments on PTB-XL \cite{wagner2020ptbxl},  SLEEP-EDF \cite{sleep_edf_expanded_v1}, and MIMIC-IV \cite{johnson2020mimiciv} dataset demonstrate that our approach yields highly competitive performance compared to existing MMFL baselines.

\end{itemize}

\section{Related Work}

\paragraph{Multimodal Learning with Missing Modalities.}
Multimodal learning has shown strong potential by integrating complementary information from multiple sources. However, most multimodal methods assume that all modalities are fully observed, which rarely holds in real-world applications. To address incomplete observations, prior studies have explored heuristic or statistical imputation, neural reconstruction models, and generative approaches that recover missing modalities from available ones \cite{zhou2022missing, chen2024probabilistic, wang2023distribution, yang2024frequency}. More recently, pretrained models have also been leveraged to improve incomplete multimodal learning \cite{das2024decoder, xue2023promptcast, cao2023tempo, rasul2023lag}. Despite their success, most of these methods are developed in centralized settings and usually assume access to all data or sufficiently complete multimodal observations during training, which conflicts with the privacy constraints of federated learning \cite{nguyen2025pepsy}.

\paragraph{Multimodal Federated Learning.}
Federated learning enables multiple clients to jointly optimize a shared model without exchanging raw data, making it particularly attractive for privacy-sensitive domains such as healthcare \cite{zhao2022multimodal, huang2024multimodal, che2023multimodal}. Recent MMFL methods mainly focus on modality fusion, shared representation learning \cite{chen2024feddat, peng2024fedmm, phung2025federated}, and disentangled feature construction \cite{xiong2022unified, yu2023multimodal}. While these methods improve collaborative learning across distributed clients, many of them still assume relatively homogeneous modality configurations or do not explicitly address severe and fine-grained missingness in multimodal data.

\paragraph{Missingness in Multimodal Federated Learning.}
A central challenge in MMFL is that missingness may arise at two levels: modality-level missingness \cite{chen2022fedmsplit, phung2024contrastive} and within-modality missingness \cite{yu2024robust, nguyen2024fedmac}. Recent frameworks like PEPSY \cite{nguyen2025pepsy} and FlexMoE \cite{yun2024flex} attempt to adapt to these patterns through reconfigurable embeddings or expert routing. However, these "implicit" methods, along with those relying on deterministic imputation like FuseMoE \cite{han2024fusemoe}, often fail to recover the underlying data distribution. Such passive approaches are fundamentally inadequate for clinical applications where missing data carry high uncertainty and strong cross-modal correlations (e.g., the link between blood pressure and heart rate \cite{zhang2001dynamic}). As noted in \cite{lincyin2026}, failing to explicitly ground these gaps can introduce semantic ambiguity and degrade expressive power.

\paragraph{Time-series Imputation and Diffusion-based Generation.}
Missing values are a long-standing challenge in multivariate time-series analysis \cite{wang2024deep, johnson2020mimiciv}. Early predictive approaches, including CNN-based \cite{wu2022timesnet}, RNN-based \cite{yoon2018estimating}, and Transformer-based models \cite{bansal2021missing}, exploit temporal dependencies for reconstruction. Generative models like GP-VAE \cite{fortuin2020gp} and GRUI-GAN \cite{luo2018multivariate} offered more flexible distributions, leading to the recent success of diffusion-based frameworks. Models such as CSDI \cite{tashiro2021csdi}, SSSD \cite{alcaraz2022diffusion}, and LSCD \cite{fons2025lscd} treat missing values as latent variables for conditional generation. Additionally, masked-modeling approaches like Timer \cite{liu2024timer} and MOMENT \cite{goswami2024moment} further demonstrate the power of self-supervised reconstruction. Unlike unimodal diffusion or signal-level models \cite{cao2024timedit, darlow2024dam}, our method incorporates cross-modal conditional diffusion to explicitly generate missing temporal observations from available multimodal context.

\begin{figure*}[!t]
    \centering

    \includegraphics[width=0.85\textwidth]{figs/structure.png} 
    \caption{{Overview of the proposed \method framework.} 
    (a){ Global Federated Workflow.} The server coordinates the distribution and aggregation of global model parameters $\Theta$ and prompt embeddings $\mathcal{W}$ across clients via FedAvg. 
    (b){ Local Two-Phase Pipeline \& Inference.} Inside each client, the process is structurally decoupled. In {Phase A}, the missing pattern ($\mathcal{W}_{cond}$) is extracted to explicitly guide the diffusion model in imputing missing data. In {Phase B}, the model optimizes the modality profile ($\mathcal{W}_{mod}$) and trains the instance encoder. During the inference stage (dotted line), the explicitly imputed data is fed back into the encoder to extract refined instance features ($\mathcal{W}_{ins}$), which are then concatenated with $\mathcal{W}_{mod}$ and $\mathcal{W}_{cond}$ for downstream task prediction.}
    \label{fig:framework}
\end{figure*}

\section{Methodology}

\subsection{Problem Statement}
\paragraph{Federated System Setup.}
We consider a Federated Multimodal Learning (MMFL) system consisting of one central server and $K$ isolated clients, indexed by $k \in \{1, 2, \dots, K\}$. Let $\mathcal{D}_k$ denote the private local dataset held by the $k$-th client, with a local sample size of $n_k = |\mathcal{D}_k|$. In real-world medical scenarios, data distributions across different clients are not strictly mutually exclusive (e.g., the same patient might have records in multiple institutions). Therefore, the global dataset is formally defined as the union of all local datasets, $\mathcal{D}_{\text{global}} = \bigcup_{k=1}^K \mathcal{D}_k$. The total number of unique samples across the entire federated system is $N_{\text{total}} = |\mathcal{D}_{\text{global}}|$, where $\sum_{k=1}^K n_k \ge N_{\text{total}}$ due to potential overlapping records. Following the standard privacy protocols of federated learning, raw data $\mathcal{D}_k$ is strictly kept local and cannot be shared across clients or uploaded to the server for explicit deduplication.

\paragraph{Modality Data Missing.}
Suppose the system involves a total of $M$ distinct modalities. For the $b$-th sample in a local dataset, we denote the input sequence of the $m$-th modality as $x_b^{(m)} \in \mathbb{R}^{L_{ts} \times L_{f}}$, where $L_{ts}$ and $L_{f}$ represent the temporal length and feature dimensionality, respectively.
In real-world federated scenarios, clients often suffer from severe modality heterogeneity and instance-level missingness (e.g., due to different sensor configurations across hospitals). To mathematically formulate this, we define a binary observability indicator $r_b^{(m)} \in \{0, 1\}$ for the $b$-th sample. Specifically: $r_b^{(m)} = 1$ indicates that the $m$-th modality is strictly observable for this sample and $r_b^{(m)} = 0$ indicates that the $m$-th modality is missing.
Consequently, the actually observed data for the $b$-th sample can be denoted as a set $\mathcal{X}_b = \{x_b^{(m)} \mid r_b^{(m)} = 1, m \in \{1, \dots, M\}\}$. The observable modality indicator set for this sample is $\mathcal{O}_b = \{i \mid r_b^{(i)} = 1\}$.

\paragraph{Within modality missing.}
Beyond the absence of entire modalities, individual modalities may also exhibit partial missingness (e.g., missing specific time steps or feature dimensions). First we resample all modalities into the same length. 
To characterize this fine-grained missingness, we introduce a modality-specific binary observation mask $R^{(m)} \in \{0, 1\}^{L_{ts} \times L_{f}}$.
Consequently, the observed and unobserved components of the representation for modality $m$ can be formally expressed as: $x_b^{(m, \mathcal{O})} = R^{(m)} \odot x_b^{(m)}$ and $x_b^{(m, \mathcal{U})} = (\mathbf{1} - R^{(m)}) \odot x_b^{(m)}$, where $\odot$ denotes the Hadamard product.
To facilitate joint multimodal processing in subsequent generative modules, these modality-specific components are concatenated along the feature dimension to form the unified sample-level observed and unobserved representations: 
\begin{equation}
x_b^{\mathcal{O}} = \text{Concat}_{m=1}^M \big( x_b^{(m, \mathcal{O})} \big),  x_b^{\mathcal{U}} = \text{Concat}_{m=1}^M \big( x_b^{(m, \mathcal{U})} \big) \text{.}
\end{equation}

\paragraph{Learning Objective.}
Our framework collaboratively optimizes a global model $\Theta$ across $K$ clients while addressing severe modality and intra-modality missingness. This is formulated via two synergistic objectives: a global federated goal and a local intermediate estimation goal.
(1) \textbf{Global Federated Objective.} 
The central system aims to find the optimal global parameters $\Theta$ by minimizing a federated empirical risk $\mathcal{L}$, which is defined as the weighted average of local losses across all participating clients:
\begin{equation}
\min_{\Theta} \mathcal{L}(\Theta) = \sum_{k=1}^K \frac{n_k}{N_{\text{total}}} \mathcal{L}_k(\Theta; \mathcal{D}_k),
\end{equation}
where $n_k$ is the local sample size and $N_{\text{total}}$ is the total number of unique samples. For each client $k$, the local loss $\mathcal{L}_k$ is evaluated using conditionally completed representations to ensure the global model captures robust cross-modal dependencies.
(2) \textbf{Local Intermediate Estimation Objective.} 
To mitigate the impact of data sparsity, each client independently optimizes an intermediate imputation objective prior to or during the global training process. Specifically, For a sample $x_b$, the client aims to model the conditional distribution $p_{\theta}(x^{\mathcal{U}}_b \mid x^{\mathcal{O}}_b, c_{\text{cond}})$ of unobserved components $x^{\mathcal{U}}_b$ given the missing pattern context $c_{\text{cond}}$.

\subsection{Overview Workflow}
We propose a federated learning framework that decouples local estimation from global task fusion. The process is structured as a coordinated cycle:
\begin{itemize}
    \item \textbf{Step 1: Global Broadcast.} At the start of round $t$, the server distributes the global parameters $\Theta^{(t)}$ and embedding matrices $\mathcal{W}^{(t)}$ to a selected subset of clients. 

    \item \textbf{Step 2: Local Multi-Phase Training.} Each client $k$ updates the model using its private data $\mathcal{D}_k$ through two sequential phases: 
    (i) \textbf{Phase A (Conditional Estimation)}, which focuses on the intermediate goal of imputing missing data using available observed modalities; and 
    (ii) \textbf{Phase B (Task Optimization)}, which optimizes the instance-specific encoder, modality-specific embedding spaces, and the classification module using available local data. Note that the explicitly imputed raw data generated by the diffusion model are subsequently fed into these trained extractors during the inference stage. 

    \item \textbf{Step 3: Parameter Upload.} Clients pack their updated local weights $\Theta_k^{(t+1)}$ and embeddings $\mathcal{W}$ for transmission back to the server.

    \item \textbf{Step 4: Global Aggregation.} The server applies Federated Averaging (FedAvg) to both the parameters $\Theta^{(t+1)}$ and embeddings $\mathcal{W_*}^{(t+1)}$ to produce the global state for the next round:
    \begin{equation}
\Theta^{(t+1)} = \sum_{k=1}^K \frac{n_k}{N_{\text{total}}} \Theta_k^{(t+1)}, \mathcal{W}_*^{(t+1)} = \sum_{k=1}^K \frac{n_k}{N_{\text{total}}} \mathcal{W}_{*,k}^{(t+1)} \text{.}
    \end{equation}
\end{itemize}

\paragraph{Definitions of Learnable Embeddings.} 
To effectively orchestrate this multi-phase pipeline and decouple the complex missing scenarios, we introduce a set of learnable prompt embeddings: $\mathcal{W} = \{\mathcal{W}_{\text{mod}}, \mathcal{W}_{\text{ins}}, \mathcal{W}_{\text{cond}}\} \in \mathbb{R}^{M \times M \times D}$, where $D$ denotes a custom-defined embedding dimension. 
In {Phase A}, the focus is on the missing pattern: $\mathcal{W}_{\text{cond}}$ is learned to serve as the explicit condition guiding the diffusion generation. Subsequently, in {Phase B}, task optimization engages the full embedding set: $\mathcal{W}_{\text{mod}}$ is optimized as a structural feature to maintain a stable modality profile, $\mathcal{W}_{\text{ins}}$ is extracted by instance encoder to encode fine-grained instance-level semantics, and $\mathcal{W}_{\text{cond}}$ is continually utilized to align with the downstream task. Finally, during the {inference stage}, $\mathcal{W}_{\text{ins}}$ is re-extracted from the explicitly imputed data to ensure strict semantic consistency. Ultimately, these embeddings are concatenated for robust downstream classification.

\subsection{Phase A: Conditional Estimation} 

\begin{figure}[t]
    \centering
    \includegraphics[width=0.45\textwidth]{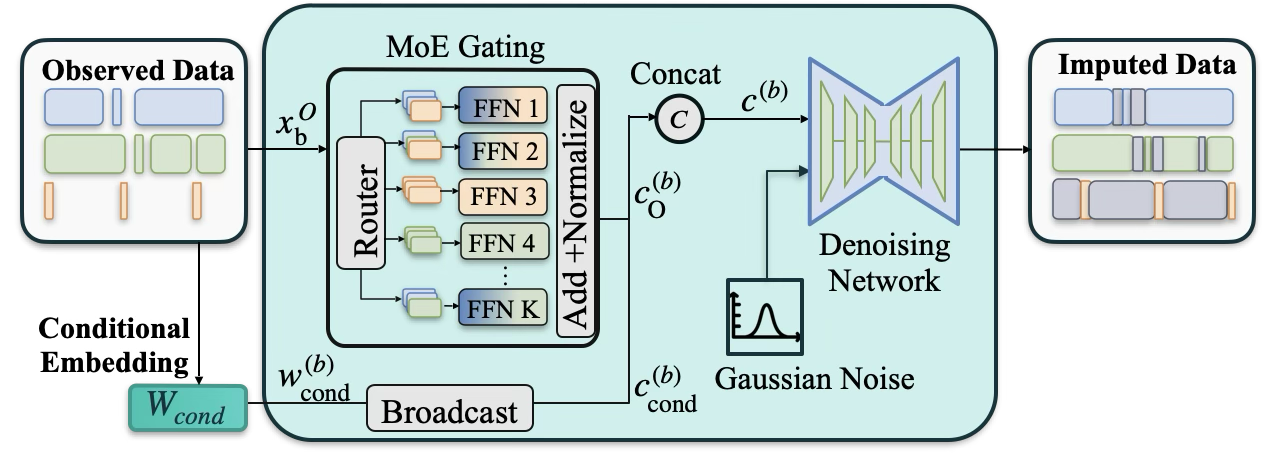}
    
    \caption{Detailed architecture of the Conditional Diffusion module. 
    The module performs fine-grained data imputation by integrating observed context and missing conditions. Specifically, available data $x_{b}^{\mathcal{O}}$ is processed via an MoE Gating block to extract observed contextual feature $\mathbf{c}_{\mathcal{O}}^{(b)}$. The cross-modal embedding $\mathcal{W}^{(m)}_{cond}$ provides structural priors regarding missing patterns. These two streams are concated to form the unified condition $\mathbf{c}^{(b)}$, which guides the Denoising Network in imputing unobserved components from Gaussian noise.}
    \label{fig:diffusion}
\end{figure}

\paragraph{Cross-Modal Conditional Embedding.}
Building on its role as a structural prior, the global embedding $\mathcal{W}_{\text{cond}} \in \mathbb{R}^{M \times M \times D}$ explicitly models the correlation rules between sensors. Formally, the element $\mathcal{W}_{\text{cond}}[i, m] \in \mathbb{R}^D$ is a parameterized vector representing the specific conditional contribution from a source modality $i$ to a target modality $m$. By parameterizing these relationships rather than relying on sample-level retrieval, $\mathcal{W}_{\text{cond}}$ is naturally suited for federated aggregation and end-to-end optimization.

For a specific sample $b$, let $\mathcal{O}_b = \{i \mid r_b^{(i)} = 1\}$ denote the set of its currently observable modalities. The core of our design is to provide a unified condition routing mechanism regardless of the missing severity. We formulate the sample-level condition vector $\mathbf{\omega}_{\text{cond}}^{(b, m)}$ as follows:
\begin{equation}
\mathbf{\omega}_{\text{cond}}^{(b, m)} =
\begin{cases}
\mathcal{W}_{\text{cond}}[m,m], & \text{if } r_b^{(m)}=1, \\
\frac{1}{|\mathcal{O}_b|}\sum_{i \in \mathcal{O}_b}\mathcal{W}_{\text{cond}}[i,m], & \text{if } r_b^{(m)}=0.
\end{cases}
\end{equation}
This design guarantees a strict and adaptive conditional logic: if a modality is completely absent $r_b^{(m)}=0$, it aggregates cross-modal knowledge from surviving modalities to maintain a structural placeholder; crucially, if the modality is partially observable $r_b^{(m)}=1$, it utilizes its robust self-condition to explicitly guide the fine-grained diffusion imputation of its intra-modality missing values $x_b^{\mathcal{U}}$.

\paragraph{Conditional Diffusion.}As shown in Figure \ref{fig:diffusion}, we inject the cross-modal condition as crucial side information into the diffusion process. For a given sample $b$, rather than relying on a single target modality, we concatenate the conditional vectors across all available modalities $m \in \mathcal{M}$ to form a comprehensive sample-level condition $\mathbf{\omega}_{\text{cond}}^{(b)} = \text{Concat}_{m \in \mathcal{M}}\big(\mathbf{\omega}_{\text{cond}}^{(b,m)}\big)$. This aggregated condition is then broadcast along the temporal dimension to form the condition sequence $c_{\text{cond}}^{(b)}$. Simultaneously, the available observed data $x_{b}^{\mathcal{O}}$ is processed by a feature encoder equipped with a Mixture-of-Experts (MoE) module \cite{shazeer2017outrageously}. This MoE structure dynamically selects and routes useful information to generate the robust observed contextual feature $c^{(b)}_{\mathcal{O}}$. Finally, these two representation streams are concatenated to form the complete multimodal condition $c^{(b)}$:
\begin{equation}
c^{(b)} = [c^{(b)}_{\mathcal{O}} 
\oplus c_{\text{cond}}^{(b)}] \text{.}
\end{equation}

We implement the estimation via a conditional denoising diffusion model. 
To distinguish diffusion latents from instance notation, we use $z_t^{(b)}$ to denote the noisy latent for sample $b$ at step $t$, defined only over the unobserved components $x_{\mathcal{U}}^{(b)}$ while keeping observed entries fixed. 
The generative reverse process recover $z_0$ conditioned on the multimodal context $c^{(b)}$ :
\begin{equation}
p_\theta(z_{0:T}^{(b)} \mid c^{(b)}) = p(z_T^{(b)}) \prod_{t=1}^{T} p_\theta(z_{t-1}^{(b)} \mid z_t^{(b)}, c^{(b)}) \text{,}
\end{equation}
where transitions $p_\theta(z_{t-1}^{(b)} \mid z_t^{(b)}, c^{(b)})$ are parameterized by a conditional denoising network $\epsilon_\theta(z_t^{(b)}, t \mid c^{(b)})$ to estimate the added noise.

\paragraph{Training Strategy for Phase A.}

Following the noise-prediction parameterization, the local intermediate objective for client $k$ is achieved by minimizing the conditional diffusion loss over its private dataset $\mathcal{D}_k$:
\begin{equation}
    \mathcal{L}_{\text{diff}}^{(k)} = \mathbb{E}_{x_b \in \mathcal{D}_k, \epsilon, t} \left[ \left\| \epsilon - \epsilon_{\theta}(z_t^{(b)}, t \mid c^{(b)}) \right\|_2^2 \right] \text{,}
\end{equation}
where $\epsilon \sim \mathcal{N}(0, I)$ and $z^{(b)}_t = \sqrt{\bar{\alpha}_t}\,z_0^{(b)} + \sqrt{1-\bar{\alpha}_t}\,\epsilon$ is obtained via the standard DDPM \cite{ho2020denoising} reparameterization.

As ground-truth missing values are unavailable during training, we adopt the self-supervised masking strategy proposed in CSDI \cite{tashiro2021csdi}. Specifically, a subset of the observed values is randomly masked and treated as pseudo-imputation targets, while the remaining observations serve as conditioning signals. The model is trained to denoise these masked targets under the diffusion process. Crucially, unlike the unimodal baseline, our approach explicitly conditions the denoising process on the missing-aware multimodal context $c^{(b)}$, thereby enabling a richer, cross-modal conditional estimation. Ultimately, the trained conditional diffusion model is utilized to recover the unobserved components from the final denoised latent $z_0^{(b)}$. These generated components are then merged with the observed data $x_b^{\mathcal{O}}$ to construct the imputed sample $\hat{x}_b$, which serves as the holistic input for the subsequent inference stage.


\subsection{Phase B: Task Optimization}
During the federated Phase B training, the downstream feature extractors and the classifier are optimized directly on the available local observations $x_b^{(\mathcal{O})}$ to ensure robust alignment. 
During the inference stage, the fully imputed data $\hat{x}_b$ generated from Phase A is seamlessly leveraged as the definitive input to extract enriched feature representations. 
\paragraph{Modality-Specific Embedding.} 
Following PEPSY\cite{nguyen2025pepsy}, we introduce a learnable modality-specific embedding that encodes the structural origin of each feature. Formally, the profile feature for sample $b$ is directly retrieved from the learnable embedding $\mathcal{W}_{mod}$:
\begin{equation}
\mathbf{\omega}^{(b, m)}_{\text{mod}} = \mathcal{W}_{\text{mod}}[m] \in \mathbb{R}^{D} \text{,}
\end{equation}
which captures the intrinsic characteristics and static domain knowledge of the observed modality $m$, independently of the specific instance content.
\paragraph{Instance-Specific Embedding.} 
In addition to modality profiles, a fundamental challenge lies in characterizing the latent structural expectations—essentially, defining what the data distribution should look like for a given sample. During this task optimization phase, we employ an instance encoder to extract fine-grained semantic representations from the available observations:
\begin{equation}
\mathbf{\omega}^{(b, m)}_{\text{ins}} = f_{\text{ins}}\big(x^{\mathcal{O}}_b\big) \in \mathbb{R}^{D} \text{,}
\end{equation}
where $f_{\text{ins}}$ is the instance feature encoder.
While the encoder is optimized on $x^{\mathcal{O}}_b$ during Phase B to align with the downstream task, this formulation is deliberately designed to overcome the limitations of conservative methods like PEPSY \cite{nguyen2025pepsy}. PEPSY permanently restricts instance representations to fragmented partial data, hindering cross-modality alignment. In contrast, our Phase B training prepares the encoder for a crucial architectural shift: during the subsequent inference stage, this same encoder will seamlessly process the explicitly imputed data $\hat{x}_b$. By ensuring the trained encoder can ultimately integrate these synthesized cues, we guarantee that the downstream network receives a holistic and coherent input, effectively bridging the gap between partial training observations and complete semantic understanding.

\paragraph{Feature Fusion and Downstream Task.} 
To guide the final diagnostic prediction, the representation for modality $m$ of sample $b$ is formed by concatenating the instance context, the observed profile, and the cross-modal condition:
\begin{equation}
\mathbf{\omega}^{(b, m)} = \big[\mathbf{\omega}_{\text{ins}}^{(b, m)} \parallel \mathbf{\omega}_{\text{mod}}^{(b, m)} \parallel \mathbf{\omega}_{\text{cond}}^{(b, m)}\big] \in \mathbb{R}^{3D} \text{.}
\end{equation}
By doing so, the classifier makes decisions while being fully aware of both the inherent modality traits and the explicitly injected missing-pattern conditions. Finally, the individual representations $\mathbf{\omega}^{(b, m)}$ across all available modalities $m \in \mathcal{M}$ are concatenated to form a comprehensive sample embedding, which is subsequently fed into the classification head. During this phase, the parameters of the feature extractors and the classifier are jointly optimized to maximize discriminative power.


\subsection{Federated Optimization of Embeddings}
$\mathcal{W}_{\text{mod}}$ and $\mathcal{W}_{\text{cond}}$ are globally shared, learnable parameter spaces. During the local two-phase training on each client, they receive gradients and are updated by both the imputation loss and the classification loss.

On the server side, $\mathcal{W}_{\text{j}}$ is aggregated using the standard FedAvg algorithm:
\begin{equation}
\mathcal{W}_{\ast}^{(t+1)} = \sum_{k=1}^{K} \frac{n_k}{\sum_{i=1}^K n_i} \mathcal{W}_{\ast, k}^{(t+1)},
\end{equation}
where $\mathcal{W}_{\ast} \in \{\mathcal{W}_{\text{mod}}, \mathcal{W}_{\text{ins}}, \mathcal{W}_{\text{cond}}\}$ represents any of the shared embedding spaces, and $n_k$ denotes the number of samples on client $k$.

This mechanism is exceptionally powerful in handling within-modality missingness, which is empirically validated by our feature reconstruction analysis in Section \ref{sec:feature_reconstruction}. If certain conditional relationships are unobservable on one client (due to specific sensor absence), they can be learned by other clients and supplemented during the server aggregation. This achieves robust cross-client conditional knowledge transfer without the overhead of synchronizing complex missing profiles.

\section{Experiments}

\subsection{Experimental Settings}

\paragraph{Datasets and Missing-data Simulation.}
We evaluate the proposed framework on three publicly available multimodal datasets: PTB-XL \cite{wagner2020ptbxl} and SLEEP-EDF \cite{sleep_edf_expanded_v1} for physiological signal classification, and MIMIC-IV \cite{johnson2020mimiciv} for clinical time-series prediction. PTB-XL is a large-scale clinical dataset with 21,799 ECG samples, 12-lead records from more than 18,000 patients. In our setup, each of the 12 ECG leads is treated as an individual modality, resulting in a 12-modality diagnostic classification task. SLEEP-EDF contains 197 whole-night polysomnographic (PSG) recordings. We utilize five continuous physiological two EEG channels (Fpz-Cz and Pz-Oz), EOG, respiratory (Resp), and EMG as distinct modalities. These signals are segmented into 30-second epochs for a standard 5-class sleep stage scoring task (Wake, N1, N2, N3, and REM). Finally, MIMIC-IV is a comprehensive electronic health record (EHR) dataset containing rich clinical information from over 35,000 ICU patients. We adopt two benchmark tasks based on the clinical observations from the first 48 hours of ICU admission. The first is 48-hour in-hospital mortality (48-IHM), a binary classification task that predicts whether the patient will die before discharge. The second is 48-hour length-of-stay (48-LOS), a binary classification task that predicts whether the patient's ICU stay will exceed 3 days. Both tasks use only the data collected within the initial 48-hour window, which is a common setup in clinical prediction benchmarks.

To simulate incomplete multimodal observations, we follow the commonly used missing-statistics protocol. Focusing on a cross-silo federated learning scenario, where institutions are a few to be trusted or involved in collaborations, the datasets are partitioned across $K=6$ clients with 3 clients randomly selected for each training round under a non-IID setting. We vary two key parameters: the sample-level missing ratio $p_s \in \{0.2, 0.8\}$, which controls the distribution of samples with at least one missing modality, and the within-modality missing probability $p_w \in \{0.2, 0.8\}$, which is the distribution of features absent within each sample in affected modality. This setting yields four experimental configurations per dataset, spanning from slightly ($p_s{=}0.2,\, p_w{=}0.2$) to severe ($p_s{=}0.8,\, p_w{=}0.8$) missing-data scenarios. The model is trained for $T=70$ communication rounds for PTB-XL and SLEEP-EDF dataset, $T=30$ communication rounds for the MIMIC-IV dataset. Both with a client participation ratio of $0.5$ per round. All experiments are conducted with a fixed training seed to ensure reproducibility, and classification accuracy is used as the evaluation metric for both tasks.

\subsection{Experimental Results}
\paragraph{PTBXL and SLEEP-EDF Datasets.}
Table~\ref{tab:results_ptbxl_edf} shows the classification accuracy on the PTBXL and SLEEP-EDF datasets. On PTBXL, \method achieves the best results under all four missing-data settings, maintaining an accuracy around 77\% regardless of missing level. Even under the most severe condition of $p_s{=}0.8$, $p_w{=}0.8$, \method still reaches 77.23\%, while FedMM drops to only 22.08\% and FedAvg stays at 67.99\%. Compared to PEPSY, which is the strongest baseline, \method outperforms it by about 2--3\% in most cases. This shows that explicitly imputing missing data through conditional diffusion is more effective than implicitly handling it with missing embeddings. On SLEEP-EDF, \method shows a similar advantage under high sample-level missingness ($p_s{=}0.8$), where it outperforms all baselines including PEPSY by about 2--4\%. When $p_s{=}0.2$, PEPSY obtains higher accuracy than \method. A possible reason is that when only a small fraction of samples have missing data, the diffusion model has fewer imputation targets to learn from, which limits its benefit. However, \method shows better consistency as within-modality missingness grows: the accuracy gap between $p_w{=}0.2$ and $p_w{=}0.8$ for \method is less than 1\%, while for PEPSY this gap reaches about 1.5--1.7\%. Overall, \method demonstrates stronger stability and resilience than the baselines, especially when data missingness becomes more serious.

\paragraph{MIMIC-IV Dataset.}
Table~\ref{tab:results_mimic_tasks} reports the results on the MIMIC-IV dataset for two clinical tasks: 48-hour in-hospital mortality (48-IHM) and length-of-stay (LOS) prediction. On the 48-IHM task, \method achieves the best F1 score in all four settings. This is important because F1 balances precision and recall, which matters in clinical scenarios where both false positives and false negatives have real consequences. For AUROC, \method also leads in most cases. In the setting where $p_s{=}0.2$ and $p_w{=}0.8$, FedAvg reaches a slightly higher AUROC of 73.35\% compared to 73.24\% for \method, but \method provides a noticeably better F1 score (49.33\% vs.\ 46.18\%). Under the most severe setting  of $p_s{=}0.8$, $p_w{=}0.8$, \method outperforms PEPSY by about 2.5\% in AUROC and 0.6\% in F1, showing that our approach remains effective even with heavy data loss.

On the LOS task, \method obtains the highest AUROC in all settings, reaching 78.32\% in three out of four cases. While FedMM achieves good F1 scores when within-modality missingness is low ($p_w{=}0.2$), its F1 drops sharply from 67.02\% to 34.50\% as $p_w$ increases to 0.8 at $p_s{=}0.8$. In contrast, \method keeps its F1 stable at around 71\% even under severe missingness. Compared to PEPSY, \method leads in AUROC by about 3\% to 4\% across all settings. These results confirm that the conditional diffusion strategy in \method can effectively preserve useful clinical information and produce reliable predictions in challenging federated settings with incomplete data.

\begin{table}[t]
\centering
\caption{Classification accuracy for \textbf{PTBXL} and \textbf{EDF} datasets under varying missing-data configurations.}
\label{tab:results_ptbxl_edf}
\footnotesize
\setlength{\tabcolsep}{4pt} 
\begin{tabular}{c|l|cc|cc}
\toprule
 &  & \multicolumn{2}{c|}{\textbf{PTBXL}} & \multicolumn{2}{c}{\textbf{EDF}} \\
\cmidrule(lr){3-4} \cmidrule(lr){5-6}
$p_s$ & \textbf{Method} & $p_w{=}0.2$ & $p_w{=}0.8$ & $p_w{=}0.2$ & $p_w{=}0.8$ \\
\midrule
\multirow{4}{*}{0.2}
 & FedMM   & 49.84 & 38.07 & 44.25 & 41.51 \\
 & FedAvg  & 73.41 & 71.50 & 41.53 & 40.50 \\
 & PEPSY   & 74.85 & 73.30 & \textbf{54.19} & \textbf{52.71} \\
 & \method & \textbf{77.38} & \textbf{77.07} & 50.14 & 49.18 \\
\midrule
\multirow{4}{*}{0.8}
 & FedMM   & 40.85 & 22.08 & 41.05 & 41.05 \\
 & FedAvg  & 63.22 & 67.99 & 39.70 & 40.22 \\
 & PEPSY   & 74.76 & 74.10 & 47.18 & 45.46 \\
 & \method & \textbf{77.38} & \textbf{77.23} & \textbf{49.27} & \textbf{49.95} \\
\bottomrule
\end{tabular}
\end{table}

\begin{table}[t]
\centering
\caption{Classification performance for the \textbf{MIMIC-IV} dataset across different tasks under varying missing-data configurations.}
\label{tab:results_mimic_tasks}
\footnotesize
\renewcommand{\arraystretch}{0.99}
\setlength{\tabcolsep}{4.5pt}
\begin{tabular}{c|c|l|cc|cc}
\toprule
 & & & \multicolumn{2}{c|}{\textbf{48-IHM}} & \multicolumn{2}{c}{\textbf{LOS}} \\
\cmidrule(lr){4-5} \cmidrule(lr){6-7}
$p_s$ & $p_w$ & \textbf{Method} & AUROC $\uparrow$ & F1 $\uparrow$ & AUROC $\uparrow$ & F1 $\uparrow$ \\
\midrule
\multirow{8}{*}{0.2} 
 & \multirow{4}{*}{0.2} 
 & FedMM   & 74.02 & 54.43 & 76.08 & \textbf{69.91} \\
 & & FedAvg  & 65.44 & 60.52 & 74.23 & 66.82 \\
 & & PEPSY   & 62.67 & 46.65 & 78.16 & 66.33 \\
 & & \method & \textbf{74.37} & \textbf{60.82} & \textbf{78.32} & 68.75 \\
\cmidrule{2-7}
 & \multirow{4}{*}{0.8} 
 & FedMM   & 67.36 & 46.18 & 74.45 & 68.41 \\
 & & FedAvg  & \textbf{73.35} & 46.18 & 73.90 & 66.47 \\
 & & PEPSY   & 65.42 & 47.79 & 73.86 & 67.30 \\
 & & \method & 73.24 & \textbf{49.33} & \textbf{77.68} & \textbf{71.35} \\
\midrule
\multirow{8}{*}{0.8} 
 & \multirow{4}{*}{0.2} 
 & FedMM   & 72.02 & 46.18 & 73.51 & \textbf{67.02} \\
 & & FedAvg  & 70.38 & 52.71 & 71.81 & 65.58 \\
 & & PEPSY   & 61.47 & 51.57 & 74.48 & 60.84 \\
 & & \method & \textbf{73.51} & \textbf{61.11} & \textbf{78.32} & 64.14 \\
\cmidrule{2-7}
 & \multirow{4}{*}{0.8} 
 & FedMM   & 63.77 & 46.18 & 71.90 & 34.50 \\
 & & FedAvg  & 52.70 & 46.18 & 71.77 & 65.71 \\
 & & PEPSY   & 67.34 & 56.81 & 73.86 & 67.30 \\
 & & \method & \textbf{69.83} & \textbf{57.39} & \textbf{77.68} & \textbf{71.35} \\
\bottomrule
\end{tabular}
\end{table}

\subsection{Ablation Study}

To verify the effectiveness of our proposed approach, we conduct an ablation study. Our goal is to test if the \textit{\method} and \textit{imputation} modules are truly necessary for handling missing data. The experimental setup consists of 6 clients, trained over 25 rounds, with 3 clients randomly selected for each training round.

Table \ref{tab:results_ablation} shows the experimental results. By removing these modules one by one, we can clearly see that both the \textit{\method} and Imputation modules are essential. The full method, shown in the \method row, consistently achieves the highest test accuracy under all missing data conditions. If we remove only the Imputation module, only the \textit{\method} module, or both modules together, the model accuracy always drops. This proves that these two modules play a key role and are indispensable for maintaining high performance with incomplete data.

\begin{table}[t]
\centering
\caption{Classification test accuracy on the \textbf{PTBXL} dataset under varying missing-data configurations (higher is better).}
\label{tab:results_ablation}
\footnotesize
\setlength{\tabcolsep}{4.5pt} 
\begin{tabular}{c|l|cc}
\toprule
$p_s$ & \textbf{Method} & $p_w{=}0.2$ & $p_w{=}0.8$ \\
\midrule
\multirow{4}{*}{0.2}
 & No Imputation \& No \method & 72.45 & \textbf{72.45} \\
 & No Imputation & 72.29 & 72.13 \\
 & No \method & 72.29 & 72.13 \\
 & \method & \textbf{72.61} & \textbf{72.45} \\
\midrule
\multirow{4}{*}{0.8}
 & No Imputation \& No \method & 72.29 & 71.66 \\
 & No Imputation & 72.13 & 71.97 \\
 & No \method & 72.61 & 72.29 \\
 & \method & \textbf{72.77} & \textbf{72.61} \\
\bottomrule
\end{tabular}
\end{table}

\subsection{Feature Reconstruction Analysis} \label{sec:feature_reconstruction}
Beyond classification accuracy, we also examine whether our imputation method can reconstruct features that are close to the ground truth. We conduct this analysis on 628 test samples from the PTB-XL dataset. For each sample, 20\% of the time steps are randomly masked to simulate sample-level missing. We extract the classifier input features under three conditions: clean data with no missing values, missing data filled with zeros and no imputation, and missing data processed through our full Imputation and $W_{\text{cond}}$ pipeline. The features are extracted from the input of the classifier, resulting in a 4608-dimensional vector for each sample. We then measure the gap between the imputed features and the clean features using two metrics. L2 distance measures the overall magnitude deviation of the reconstructed feature vector from the ground truth. Cosine distance measures the directional deviation, reflecting whether the imputed features still point toward the correct diagnostic category.

\begin{figure}
    \centering
    \includegraphics[width=0.90\linewidth]{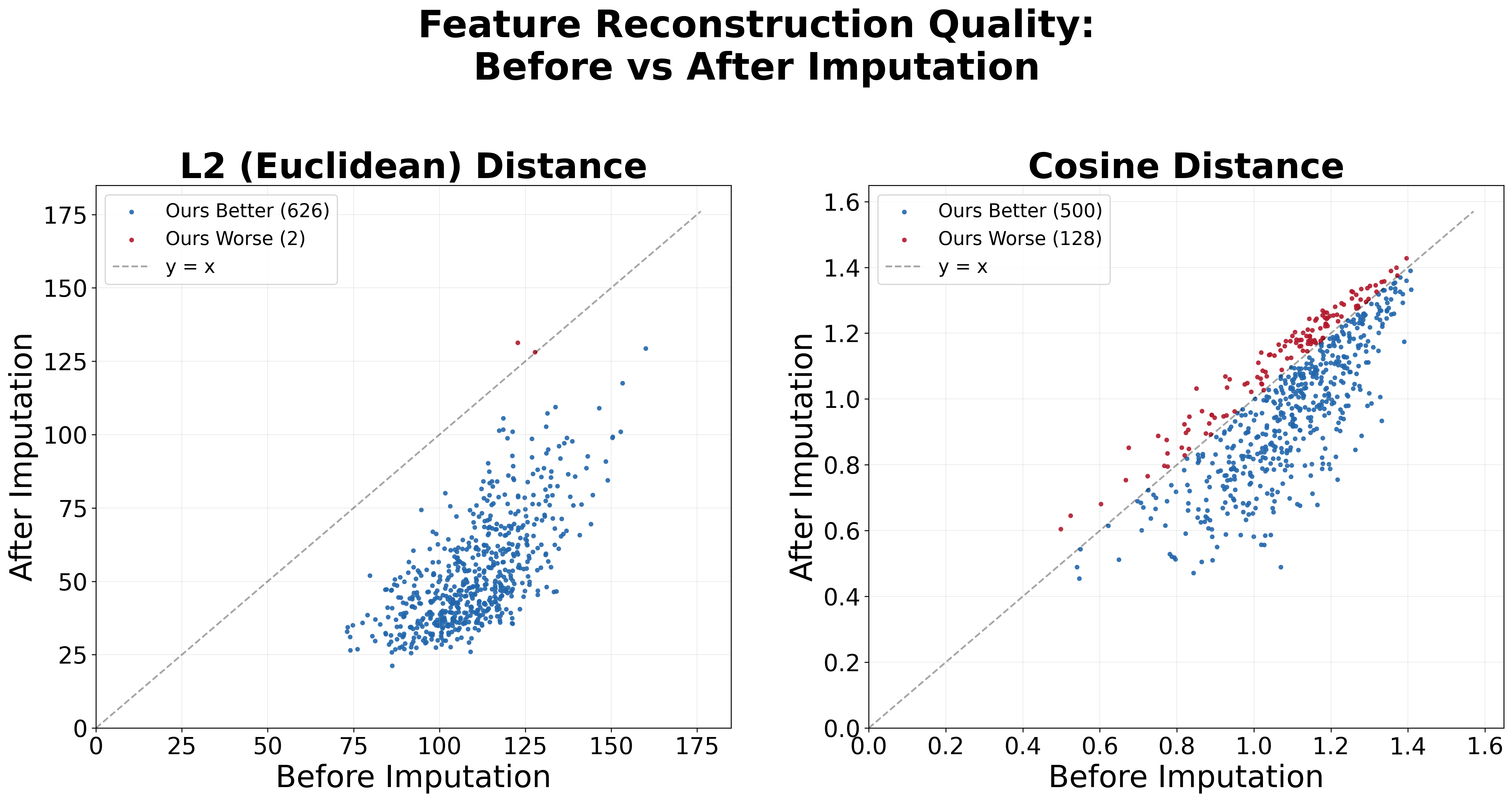}
    \caption{Per-sample feature reconstruction quality. Each point corresponds to one test sample under the within-modality missing. The x-axis measures the distance between the zero-filled feature and the ground-truth from complete data; the y-axis measures the distance after applying our conditional diffusion imputation with $W_{\text{cond}}$. Blue points fall below the diagonal $y=x$, indicating that our method produces features closer to the ground truth; red points fall above, indicating degradation. Left: L2 (Euclidean) distance. Right: Cosine distance.}
    \label{fig:imputation_scatter}
\end{figure}

As shown in Figure~\ref{fig:imputation_scatter}, most points fall below the diagonal line, meaning our method produces features closer to the ground truth than simple zero-filling. In terms of L2 distance, over 99\% of samples are improved after imputation. The cosine distance results also show a clear advantage, with about 76\% of samples benefiting from our pipeline. This confirms that the Imputation module combined with $W_{\text{cond}}$ does not merely help classification indirectly, but actively reconstructs features that resemble those obtained from complete data.

\section{Conclusion}

In this paper, we study the problem of within-modality missingness in multimodal federated learning. Existing methods handle this issue implicitly by using missing embeddings or feature alignment, which may fail to recover the true data distribution. To address this, we propose \method, a two-phase federated framework that uses conditional diffusion models to impute missing data explicitly. In Phase~A, a cross-modal conditional embedding $\mathcal{W}_{\text{cond}}$ guides the diffusion model to generate the unobserved parts based on available multimodal context. In Phase~B, modality-specific encoders and a classifier are trained for the downstream task. During inference, the imputed data is fed into the trained encoders to produce complete representations. We evaluate \method on three clinical datasets: PTB-XL, SLEEP-EDF, and MIMIC-IV. Results show that \method achieves competitive or better performance than existing baselines, especially under severe missingness. The feature reconstruction analysis further shows that our method produces features closer to those from complete data, confirming the effectiveness of the $\mathcal{W}_{\text{cond}}$ and imputation modules.

{
    \small
    \bibliographystyle{ieeenat_fullname}
    \bibliography{main}
}


\end{document}